 \documentclass[smallabstract,smallcaptions]{dccpaper}

\usepackage{graphicx} 
\usepackage{epsfig}
\usepackage{citesort}
\usepackage{cite}
\usepackage{amsmath}
\usepackage{amssymb}
\usepackage{color}
\usepackage{url}
\usepackage{multirow}
\usepackage{times}  
\usepackage{helvet}  
\usepackage{courier}  
\usepackage{booktabs}
\usepackage{xcolor}
\usepackage{array}

\usepackage{tabularx}
\usepackage{multirow}
\usepackage{threeparttable}
\usepackage{booktabs}
\usepackage{diagbox}
\usepackage{amsmath}
\usepackage{amsfonts} 
\usepackage{algorithm}
\usepackage{algorithmic}
\usepackage{threeparttable}
\usepackage{caption} 
\newlength{\figurewidth}
\newlength{\smallfigurewidth}

\begin{document}

\title
{\large
\textbf{LL-ICM: Image Compression for Low-level \\ Machine Vision via Large Vision-Language Model}
}

\author{%
Yuan Xue$^{\dagger}$, Qi Zhang$^{\ddagger}$, Chuanmin Jia$^{\sharp}$, and Shiqi Wang$^{\dagger}$\\[0.5em]
{\small\begin{minipage}{\linewidth}
\begin{tabular}{ccc}
$^{\dagger}$ School of Computer Science, City University of Hong Kong, Hong Kong SAR, China \\
$^{\ddagger}$ National Engineering Research Center of Visual Technology, Peking University, Beijing, China \\
$^{\sharp}$ Wangxuan Institute of Computer Technology, Peking University, Beijing, China
\end{tabular}
\end{minipage}}
}

\maketitle
\thispagestyle{empty}

\begin{abstract}
Image Compression for Machines (ICM) aims to compress images for machine vision tasks rather than human viewing. Current works predominantly concentrate on high-level tasks like object detection and semantic segmentation. However, the quality of original images is usually not guaranteed in the real world, leading to even worse perceptual quality or downstream task performance after compression. Low-level (LL) machine vision models, like image restoration models, can help improve such quality, and thereby their compression requirements should also be considered. In this paper, we propose a pioneered ICM framework for LL machine vision tasks, namely LL-ICM. By jointly optimizing compression and LL tasks, the proposed LL-ICM not only enriches its encoding ability in generalizing to versatile LL tasks but also optimizes the processing ability of down-stream LL task models, achieving mutual adaptation for image codecs and LL task models. Furthermore, we integrate large-scale vision-language models into the LL-ICM framework to generate more universal and distortion-robust feature embeddings for LL vision tasks. Therefore, one LL-ICM codec can generalize to multiple tasks. We establish a solid benchmark to evaluate LL-ICM, which includes extensive objective experiments by using both full and no-reference image quality assessments. Experimental results show that LL-ICM can achieve 22.65\% BD-rate reductions over the state-of-the-art methods. 
\end{abstract}

\Section{Introduction}
Image Coding for Machines (ICM) has become an emerging research topic that combines visual signal compression and understanding. Existing methods can be categorized according to the number of generated bitstreams: a single versatile bitstream for multiple tasks jointly, two bitstreams for human perception and machine task respectively and multiple bitstreams for diverse tasks separately. These ICM bitstreams can be compact representations of both original image \cite{chen2023transtic} and extracted deep features \cite{chen2023residual}, which usually have different levels of capability on texture reconstruction and semantic preservation. However, all current methods are limited by only optimizing for high-level (HL) vision tasks, while neglecting practical compression requirements for low-level (LL) ones.

Low-level vision tasks, such as denoising, debluring, inpainting, \textit{etc.}, have been studied for decades. The main purpose of these tasks is to recover the details introduced by image capturing and delivering, and enhance the visual quality of the image content. Early works tend to propose one neural network for one specific LL vision task. Recently, considering the similarity of these tasks on pixel-level content analysis and processing, more and more works are trying to create a single model to solve diverse LL vision tasks simultaneously. By introducing large vision-language modeling, state-of-the-art all-in-one LL vision models can achieve impressive performance on important LL vision tasks. These methods are also more suitable for real-world applications \textbf{\textit{where distortion types unknown or combined.}}
\begin{figure*}[t]
	\begin{center}
		\noindent
		\includegraphics[width = 6.3 in]{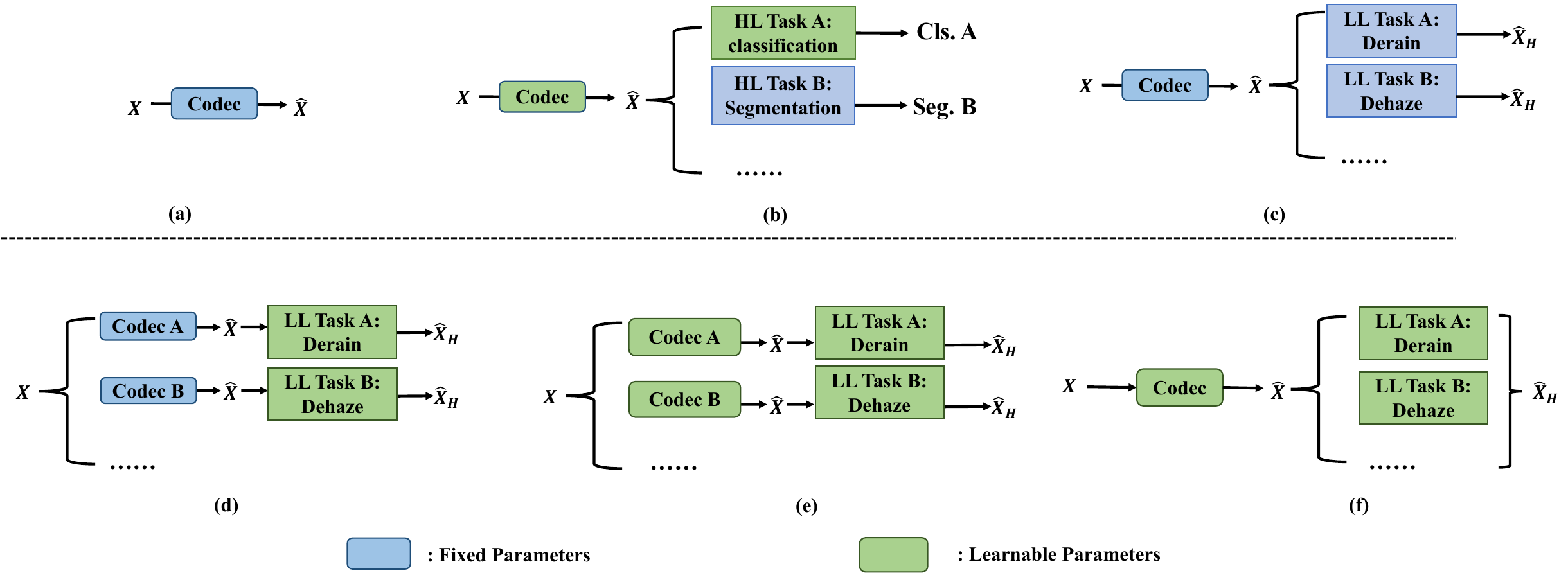}
		\caption{
  ICM frameworks from the simplest to the proposed. $\mathbf{X}$ is the original image, $\mathbf{\hat{X}}$ is the compressed image, and $\mathbf{\hat{X}_H}$ is the enhanced image by downstream task model. $\mathbf{Cls.A}$ and $\mathbf{Seg.B}$ are the results of HL tasks A and B.
    }\label{low-level SOTA framework}
	\end{center}
\end{figure*}

Image compression is usually the subsequent processing step after image capturing, which is essential for content delivery with diverse network conditions. However, compression also leads to information lost and quality degradation. Since LL visual enhancement cannot always be performed before compression due to diverse restrictions, there are two approaches to improve its performance. First, we can design LL vision models that consider compression distortions or even have the capability of removing them. But this approach has two notable drawbacks: 1) There are many types of compression distortions from different codecs like JPEG, VVC, AVS, neural codecs, \textit{etc.}, making such model hard to architect and train. 2) Without leveraging the power of LL vision models, the compression efficiency is suboptimal. For example, some pixels could have been compressed more severely and restored or enhanced appropriately in the downstream. Similarly, some signals are visually unpleasant or broken, which should not be preserved after compression. Second, and more ideally, we can create an end-to-end compression and LL vision model, which jointly optimizes the performance of enhancing the visual perceptual quality and saving the bit-rate cost at the same time.

In this work, we propose the first ICM framework for LL vision tasks, namely LL-ICM. Our goal is to achieve the rate-perception optimum by \textbf{\textit{breaking the long lasting isolation between compression and LL visual enhancement via a unified model. }} Considering the diversity and complexity of distortion types in the real world, it is expensive or even impractical to customize LL-ICM model for each individually. Therefore, we further extend our framework so that it can adapt to versatile LL vision tasks using a single all-in-one model. Specifically, we introduce large-scale vision-language model for generalized feature embedding generation and exploit diffusion procedure to remove distortions efficiently. 
Our primary contributions are delineated as follows.

\begin{itemize}


\item We propose LL-ICM, the first compression framework for LL vision. By jointly optimizing compression and LL vision processing, LL-ICM can improve the performance of both compared to existing frameworks that only consider compression or LL task alone.

\item We integrate large vision-language models (VLM) into LL-ICM. Based on the generalized feature representations learned by these models, we can use a diffusion model to reduce arbitrary type of distortions and increase the perceptual quality of compressed images. Therefore, one LL-ICM model can handle multiple LL vision tasks.

\item We create a large-scale benchmark to evaluate the performance of LL-ICM through objective evaluations. Experimental results indicate that LL-ICM significantly outperforms existing compression methods on both full and no-reference quality measurements. Specifically, LL-ICM achieves 22.65\% BD-rate reductions compared to state-of-the-art neural codecs.

\end{itemize}

\Section{Method}

\SubSection{Exsiting ICM frameworks}
We first give a brief introduction to the existing ICM frameworks and analyze their drawbacks. Fig. 1(a) is a conventional image codec optimized for signal fidelity without considering the performance of downstream tasks on compression outputs. As machine intelligence evolves, the compression needs for high-level visual understanding are getting more and more attention, leading to the establishment of HL-ICM framework shown in Fig. 1(b). In this framework, the codec is optimized for HL task models, whether their weights are fixed or not.

Most ICM methods focus on HL vision tasks. However, LL tasks are also important for applications in the real world. The reason is that the quality of original images is usually not guaranteed. After compressed and transmitted, we need to enhance their quality by LL vision processing. The simplest ICM framework for LL-vision tasks is illustrated in Fig. 1(c), where the compressed image is processed by LL task models, but LL vision models do not consider the compression artifacts. In recent years, this issue has been addressed by the framework shown in Fig. 1(d), where LL vision models are trained with adaption to the compression distortion. However, the compression efficiency is neglected. Therefore, the LL-ICM framework in Fig. 1(e) offers a better solution that jointly improves the performance of compression and LL vision processing. Nevertheless, there are many LL vision tasks and models, and it is impractical to train a codec for each of them. In this paper, we take a step further to propose a unified one-for-all LL-ICM framework to optimize the codec for diverse LL vision tasks, which is shown in Fig. 1(f).

\begin{figure}[htbp]
	\begin{center}
		\noindent
		\includegraphics[width = 5 in]{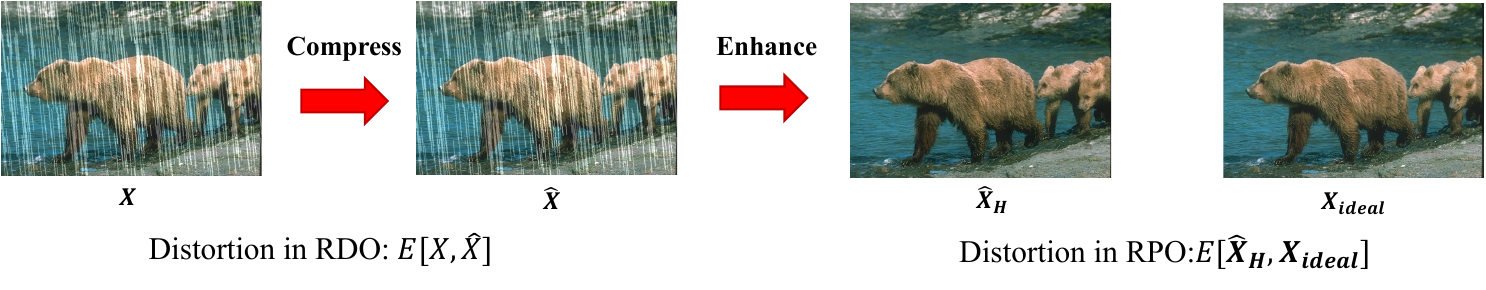}
		\caption{Difference between RPO and RDO Definition: RDO distortion measures the error of compressed image $\mathbf{\hat{X}}$ and the original image $\mathbf{X}$, while RPO measures the error of the generated enhanced image $\mathbf{\hat{X}_H}$ and the high-quality ideal image  $\mathbf{X_{ideal}}$.}\label{RPO Problem}
	\end{center}
\end{figure}

\SubSection{Problem Definition}
LL-ICM aims to jointly optimize the performance of multiple LL vision tasks while minimizing compression costs. Inherently, LL vision tasks target at removing the artifacts from images causing quality degradations and improving the perceptual quality. Therefore, the objective of LL-ICM can be defined as rate-perception optimization (RPO), which is similar to the classical rate-distortion optimization (RDO) problem in image compression. There are several differences between them. Most importantly, RDO progress assumes that the original image has an ideal level of quality, so the compression is optimized to preserve signal fidelity. However, RPO progress admits that the original image can have several flaws, so the compression should be optimized to increase the perceptual quality with the assistance of LL vision models.

As illustrated in Fig. \ref{RPO Problem}, the distortion in RDO, depends on the degradation of the original image $\mathbf{X}$, can be formulated as $\mathbb{E}(\mathbf{X}, \mathbf{\hat{X}})$, where $\mathbf{\hat{X}}$ is the compressed image, $\mathbb{E}$ is the error assessment function of compressed and original image. However, the perceptual distortion in RPO, depends on the enhancement of $\mathbf{\hat{X}}$, can be formulated as $\mathbb{E}(\tau(\mathbf{\hat{X}}), \mathbf{X_{ideal}})$. We define $\mathbf{X_{ideal}}$ as the ideal high-quality image, and $\tau(\cdot)$ is used to enhance $\mathbf{\hat{X}}$ so that its quality can be close to $\mathbf{X_{ideal}}$, and $\mathbf{\hat{X}_H}$ is the enhanced version of compressed image $\mathbf{\hat{X}}$: 
Generally, the RPO function $\mathbf{p}_{\text {opt}}$ can be described as follows:
\begin{equation}
\mathbf{p}_{\text {opt}}=\arg \min _{\mathbf{p}}\{\mathbb{E}(\mathbf{\hat{X}_H}, \mathbf{X_{ideal}})+\lambda \mathcal{R}(\mathbf{X})\},
\end{equation}
where $\mathcal{R}$ represents the bit-rate of compressed image. 


\begin{figure*}[htbp]
	\begin{center}
		\noindent
		\includegraphics[width = 6 in]{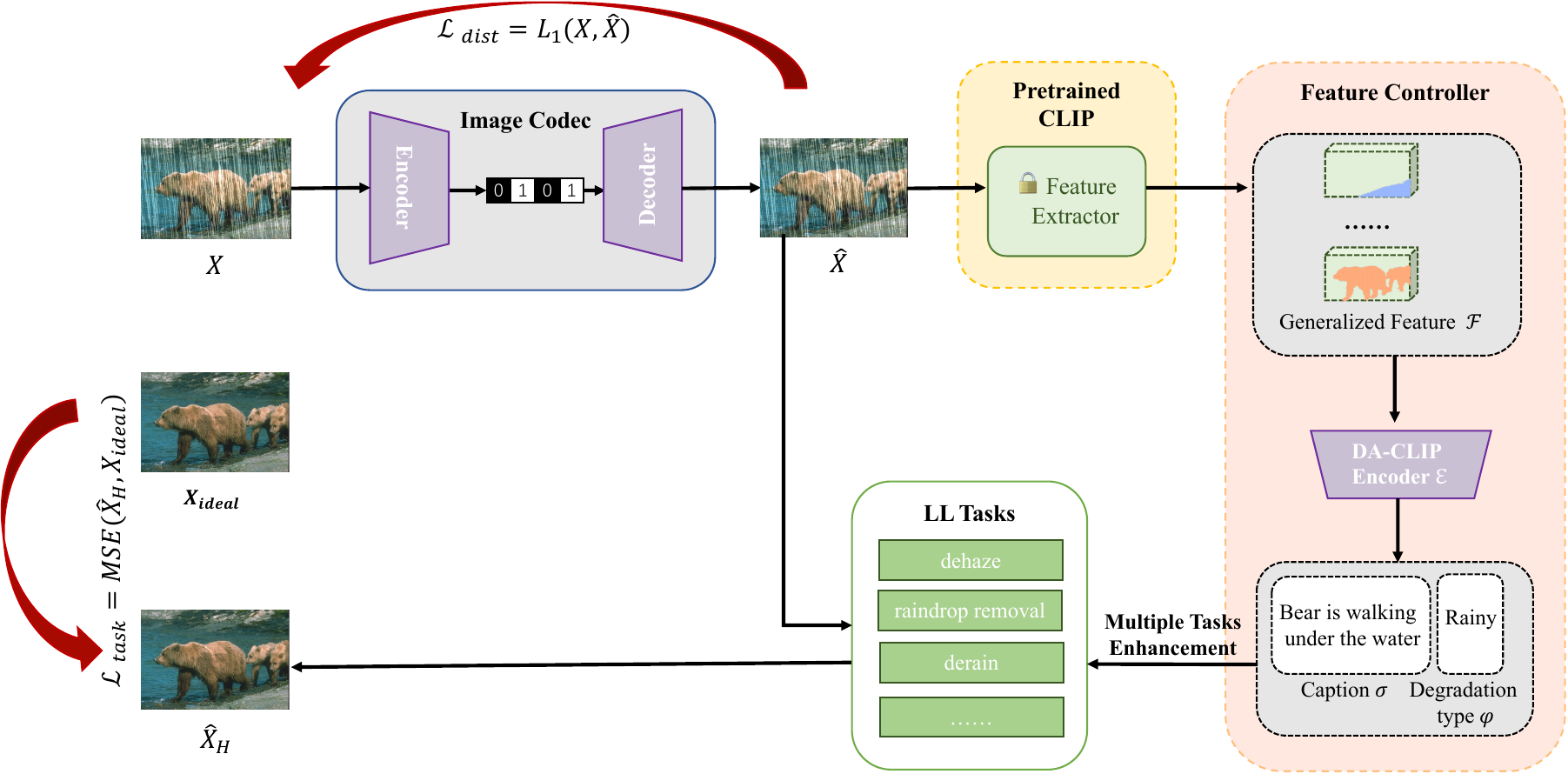}
		\caption{Overview of the LL-ICM framework: $\mathbf{\hat{X}}$ is the compressed image. A pre-trained CLIP model extracts the generalized feature $\mathbf{F}$ from $\mathbf{\hat{X}}$. After the DA-CLIP encoder $\mathbf{E}$, the encoded $\mathbf{F}$ can be used for multiple LL tasks and reconstruct the enhanced image $\mathbf{\hat{X}_H}$.}\label{LL-ICM}
	\end{center}
\end{figure*}

\SubSection{LL-ICM framework}


We present a unified LL-ICM framework in Fig. \ref{LL-ICM}, which integrates a neural image codec and a unified LL vision processing model. Notably, we incorporate a pre-trained VLM model into this framework to extract generalized features for handling different LL vision tasks. 

In this study, we employ MLIC \cite{jiang2023mlic} as backbone image codec $\mathcal{C}$ due to its superior performance. The frozen VLM model $\mathcal{V}$ \cite{radford2021learning} is used to extract generalized features $\mathbf{F}$ as:

\begin{equation}
\mathbf{F} = \mathcal{V}(\mathbf{\hat{X}})=\mathcal{V}(\mathcal{C} (\mathbf{X})).
\end{equation}

After that, the LL-vision encoder recieve $\mathbf{\hat{X}}$ and encoded $\mathbf{F}$ by encoder $\mathcal{E}$ as input reference to generate $\mathbf{\hat{X}_H}$:

\begin{equation}
\mathbf{\hat{X}_H} =\tau(\mathbf{\hat{X}}, (\mathcal{E}(\mathbf{F})).
\end{equation}

We adopt the image controller in DA-CLIP \cite{luo2023controlling} as the encoder in the feature controller of our framework, which encodes the feature $\mathbf{F}$ to LL task type $\varphi$ and caption $\sigma$. For instance, a rainy image in Fig. \ref{LL-ICM} as input of DA-CLIP can be used to extract distortion type of ``rainy'' and a caption of the image as ``bear is walking under the water''. Thus, we have:

\begin{equation}
(\varphi, \sigma) = \mathcal{E}(\mathbf{F}).
\end{equation}



This textual information, $\varphi$ and $\sigma$, guides the LL-vision enhancement network to generate $\mathbf{\hat{X}_H}$. We employ the IR-SDE \cite{luo2023image} diffusion network to handle multiple LL tasks and generate the $\mathbf{\hat{X}_H}$.



Regarding learning objective of the proposed LL-ICM $\mathcal{L}$, we aim to optimize the distortion loss $\mathcal{L}_{dist}$ between $\mathbf{X}$ and $\mathbf{\hat{X}}$, the rate loss $\mathcal{L}_{r}$, and the loss of the downstream LL task $\mathcal{L}_{task}$ between $\mathbf{\hat{X}_H}$ and $\mathbf{X_{ideal}}$. The total loss $\mathcal{L}$ of the LL-ICM can be formulated as:

\begin{equation}
\label{Loss}
\mathcal{L} = \alpha \cdot \mathcal{L}_{dist} + \beta \cdot \mathcal{L}_{r} + \gamma \cdot \mathcal{L}_{task},
\end{equation}
where the MSE loss are calculated for $\mathcal{L}{dist}$, and $\mathcal{L}{r}$ represents the bitrate of $\mathbf{\hat{X}}$. The $\mathcal{L}{task}$ is the diffusion loss, as described in IR-SDE \cite{luo2023image}, which ensures that $\mathbf{\hat{X}_H}$ is as close as possible to $\mathbf{X_{ideal}}$.


\Section{Experiment}
\SubSection{Training Setting}


Our training procedure is divided into two stages. In the first stage, we only train the image codec to achieve high compression efficiency. We train the codec on two NVIDIA 4090 GPUs for 30 epochs, using the Adam optimizer with a batch size of 16. We randomly select $2\times10^5$ images from the COCO2017 and ImageNet datasets as the initial training set, and the images are randomly cropped to a patch size of 448x448. Following the settings of CompressAI library, we set $\beta \in \left \{0.0002, 0.0008, 0.0018, 0.0035, 0.0130, 0.0350\right \}$ for $\mathcal{L}_{dist}$. The parameters $\alpha$ and $\gamma$ are set to be 1 and 0, respectively.

In the second stage, we load our pre-trained codec from the first stage. The LL task is then trained jointly with the codec and optimize it jointly for compression and LL vision tasks. In this work, we choose \textbf{\textit{dehazing, raindrop removal, deraining, deshadowing, denoising, and inpainting}} as exemplar LL vision tasks. It should be noticed that the proposed LL-ICM framework can be extended to even more tasks because of the integrated large vision-language model. To further increase the generalizability of our models, we incorporate several in-the-wild datasets into our training, such as LOL \cite{wei2018deep} and RESIDE-6k \cite{qin2020ffa}. The training datasets in this stage are listed in TABLE \ref{tab: dataset}. Noticeably, we manually generate several distortions on datasets of the denoising task by adding gaussian noise to the original images with a level of 50. 

\begin{table}[h!]
\centering
\caption{\centering{\scshape Training and Testing Datasets Used for Evaluating LL-ICM}}
\label{tab: dataset}
\fontsize{8}{8.5}\selectfont  
\begin{tabular}{c|c|c|c|c|c}
\toprule
\textbf{LL Task Type} & \textbf{Dataset} & \textbf{Size(Train+Test)} & \textbf{LL Task Type} & \textbf{Dataset} & \textbf{Size(Train+Test)} \\ [0.5ex]
\hline\hline

deraining          & Rain100H \cite{yang2017deep} & 1899+100 & dehazing        & RESIDE-6k \cite{qin2020ffa} & 6000+1000 \\
\hline

\multirow{4}{*}{inpainting} &  & \multirow{4}{*}{29901+100} &\multirow{4}{*}{denoising} & DIV2K & \multirow{4}{*}{3550+68}\\
& CelebaHQ-256 \cite{lugmayr2022repaint} & & & denoising &\\
& RePaint \cite{lugmayr2022repaint} &  & & Flick2k & \\
& & && CBSD68\cite{martin2001database} &\\
\hline

deshadowing & SRD \cite{qu2017deshadownet} & 2680+408 &raindrop removal & Raindrop \cite{qian2018attentive} & 861+58\\
\hline


\end{tabular}
\end{table}








\SubSection{Testing Setting}
To evaluate the performance of our proposed LL-ICM framework, we test the compression and LL vision task performance and compare our method with several state-of-the-art image codecs. The selected codecs include both traditional block-based codec like Enhance Enhanced Compression Model (ECM) \cite{abdoli2024video}, and deep learning-based ones like Balle2018\cite{balle2018variational}, Cheng2020 \cite{cheng2020learned}, and MLIC \cite{jiang2023mlic}.
To ensure the fairness of the comparison, we also use IR-SDE \cite{luo2023image} as the LL vision model to generate enhance outputs of the compressed images from these codecs. We conduct objective experiments by calculating both full-reference and no-reference image quality metrics on the generated $\mathbf{\hat{X}_H}$.

Specifically, we use the popular full-reference image quality assessment (FR-IQA) metric LPIPS \cite{zhang2018unreasonable} and the state-of-the-art no-reference image quality assessment (NR-IQA), Q-Align \cite{wu2023q} and LIQE \cite{zhang2023blind}, in objective test. As FR-IQA, we use the image pairs in the datasets as $\mathbf{X_{ideal}}$ and $\mathbf{\hat{X}_H}$, respectively. In contrast, we evaluate only the generated $\mathbf{\hat{X}_H}$ in NR-IQA without using $\mathbf{X_{ideal}}$ as the reference. Here, we introduce NR-IQA metrics because $\mathbf{X_{ideal}}$ is often inaccessible or even does not exist in the real world.
We also calculate the average coding gains and perceptual quality enhancements using the Bjontegaard Delta-rate (BD-rate) metrics. The BD-LPIPS and BD-Q-Align metrics are computed similarly to BD-PSNR in \cite{bjontegaard2001calculation} by substituting the distortion metric with LPIPS or Q-Align. A negative BD-rate value indicates rate savings at equivalent quality levels. Negative BD-LPIPS and positive BD-Q-Align values signify quality improvements at the same bit rate.

\begin{figure*}[htpb]
	\begin{center}
		\noindent
		\includegraphics[width = 6. in]{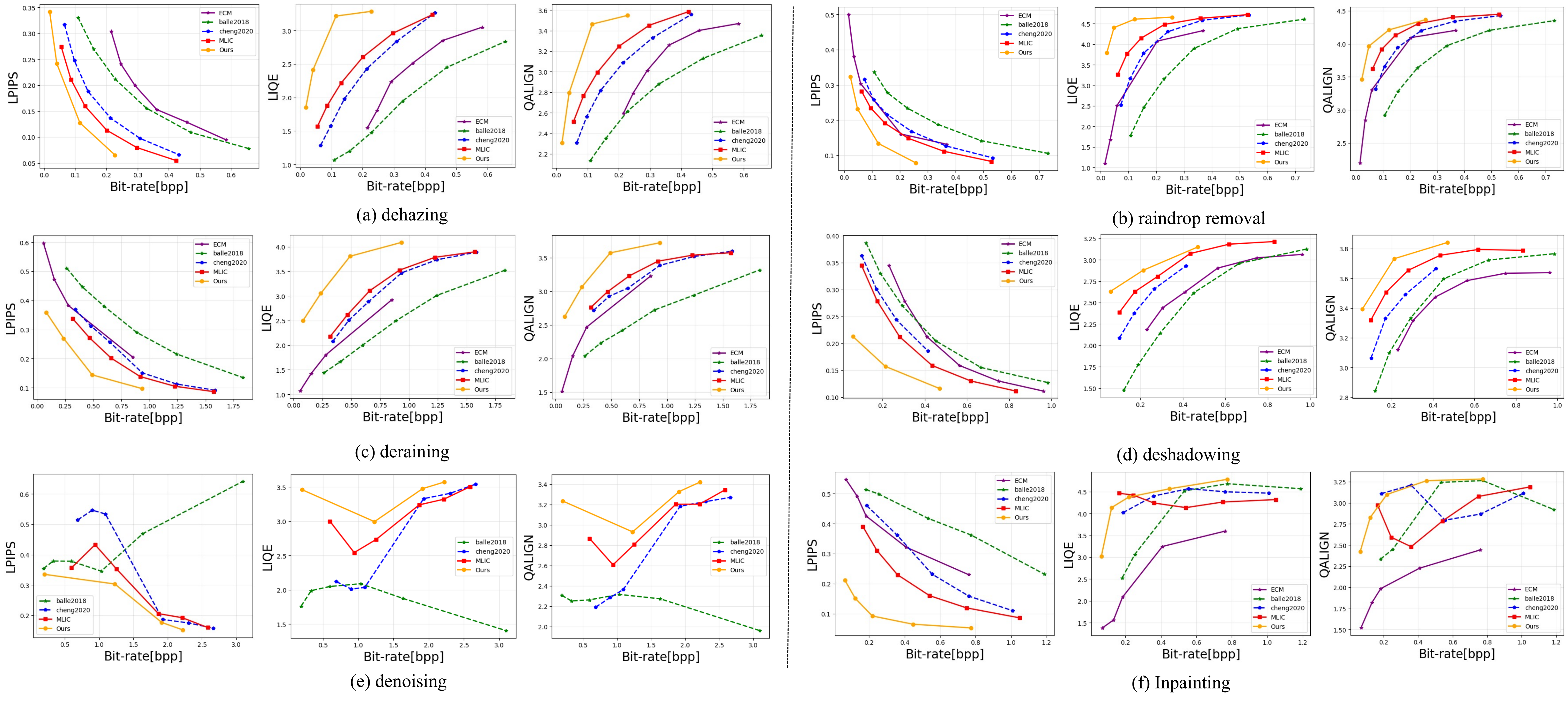}
		\caption{RP performance on 6 LL tasks: Our LL-ICM framework significantly outperforms all existing codecs. RP performance on (a)-(d) are the monotonic cases in which the perception quality increases as the bpp increases. The (e) and (f) are abnormal cases as the perception quality does not follow the regular law as (a)-(d).}\label{all rd curve}
	\end{center}
\end{figure*}

\SubSection{Rate-percetion Performance Result}
Fig. \ref{all rd curve} illustrates the rate-perception (RP) performance results on different tasks. Fig. \ref{all rd curve} (a)-(d) represents the results on dehazing, Raindrop removal, deraining, and deshadowing, respectively. 
The results show the average quality and bitrate in different datasets. It is easy to observe that the perceptual quality of all anchors in Fig. \ref{all rd curve} (a)-(d) increases as the bpp increases. Our LL-ICM model achieves the best performance among the four tasks. To be specific, the LPIPS keeps the lowest and Q-Align situated at the top of the diagram.

\begin{figure*}[htbp]
	\begin{center}
		\noindent
		\includegraphics[width = 6 in]{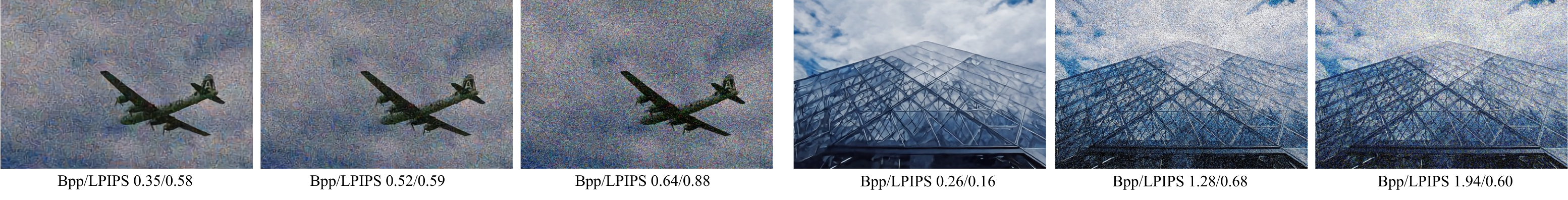}
		\caption{An abnormal result on denoising: the noise information is more in higher bpp image compared to lower bpp image.}\label{abnormal}
	\end{center}
\end{figure*}

However, not all LL tasks keep such regular results. Fig. \ref{all rd curve} (g) shows the results of de-noise. As shown in the anchor “Balle2018”, the perception performance abnormally decreases as the bpp increases. We visualize an example in Fig. \ref{abnormal}. The noisy information is even more in $\mathbf{\hat{X}_H}$ with higher bpp. Similar results are also shown in our LL-LCM framework. A turning point is shown in the Q-Align/LIQE diagram, representing the enhancement quality increases when bpp increases after the turning point. Thus, we can conclude that as the bpp increases, the codec keeps more detail and chaotic signal in CLQ, which will interfere with the de-noise downstream. Fortunately, our LL-ICM framework still performs best among the anchors. Similarly, the trend is shown in Fig. \ref{all rd curve} (h).

Interestingly, not all LL tasks yield consistent results. Fig. \ref{all rd curve} (e) shows the results for the denoising task. As shown in the result from method ``Balle2018'', the performance abnormally decreases as the bpp increases. We visualize a representative example in Fig. \ref{abnormal}, where the noisy signal in $\mathbf{\hat{X}_H}$ actually increases with higher bpp. Similar results are observed in our LL-ICM framework. A turning point is evident in the Q-Align diagram, indicating that the quality improves when the bpp increases beyond this point. Therefore, we can conclude that as bpp increases, the codec retains more detail and chaotic signals in $\mathbf{\hat{X}}$, which can interfere with the denoising process. Nevertheless, our LL-ICM framework still outperforms other methods in this extreme condition. Besides, the turning point in MLIC is also obvious as shown in Fig. \ref{all rd curve} (f). The quality decreases, and the inpainting result becomes worse when the bpp increases before this point.

\bgroup
\def\arraystretch{1}{
    \begin{table}[htpb]
    \belowrulesep=0pt
    \aboverulesep=0pt
    \centering
    \caption{\centering{\scshape BD-Rate, BD-LPIPS, and BD-Qalign results for different LL tasks with Balle2018\cite{balle2018variational} being the anchor}}
    \resizebox{1\columnwidth}{!}{
        \begin{tabular}{cc|ccc|ccc|ccc}
        
        \toprule[0.5mm] &\multirow{2}{*}{Tasks} &\multicolumn{3}{c}{dehazing} &\multicolumn{3}{c}{raindrop removal} &\multicolumn{3}{c}{denoising} \\
        \cmidrule{3-11} & 
        & \text { BD-Rate (\%) } $\downarrow $
        & \text { BD-LPIPS} $\downarrow $
        & \text { BD-QAlign} $\uparrow $
        
        & \text { BD-Rate (\%) } $\downarrow$
        & \text { BD-LPIPS} $\downarrow $
        & \text { BD-QAlign} $\uparrow$ 

        & \text { BD-Rate (\%) } $\downarrow$
        & \text { BD-LPIPS} $\downarrow $
        & \text { BD-QAlign} $\uparrow$ \\

      \midrule

      &Cheng2020  &-45.22 &-0.08 &0.51 &-40.07 &-0.06 &0.49 &-53.36 &-0.05 &0.45\\
      &ECM&-52.07 &-0.10 &0.35  & -55.02 &-0.08 &0.55 &-49.13 &-0.02 &0.43 \\
      &MLIC  &-59.72 &-0.11 &0.69 & -54.37 &-0.08 &0.63 &  -91.89 &-0.10 &0.63\\
      \cmidrule{1-11}
      &LL-ICM (Ours)  &  \textbf{-76.38} & \textbf{-0.17} & \textbf{1.18}& \textbf{-78.72} & \textbf{-0.16} & \textbf{0.93} &\textbf{-79.79} & \textbf{-0.03} & \textbf{0.43}  \\
      \midrule

      & Tasks & \multicolumn{3}{c}{deraining} & \multicolumn{3}{c}{deshadowing}   &\multicolumn{3}{c}{inpainting}\\
      \midrule
      &Cheng2020 & -41.53  & -0.11 & 0.58 &-25.32 &-0.04 &0.23  &-52.65 &-0.13 & 0.14\\
      &ECM & -48.96  & -0.12 & 0.52 &-15.07 &-0.01 &0.13 &-55.44 & -0.12 &-0.63  \\
      &MLIC & -51.46  & -0.14  & 0.66 &-35.53 &-0.05 &0.28  &-73.17 &0.22 &-0.15 \\
      \cmidrule{1-11}
      &LL-ICM (Ours) & \textbf{-77.17} & \textbf{-0.25} & \textbf{1.18}  &\textbf{-74.08} & \textbf{-0.14} & \textbf{0.51} &\textbf{-96.07} & \textbf{-0.38} & \textbf{0.36}\\
      \midrule


      \bottomrule[0.5mm]
    \end{tabular}
    }
    \label{Bjontegaard metrics}
\end{table}
}
\egroup



TABLE \ref{Bjontegaard metrics} summarizes the average coding gains and perceptual quality score enhancements on all evaluated LL tasks. LL-ICM outperforms all compared methods. The results demonstrate a substantial increase in perceptual quality score, with LL-ICM \textbf{\textit{achieving 7.70\%-49.23\% coding gains}} across various LL tasks compared to state-of-the-art methods. At the same time, the LPIPS score is significantly improved by 0.06-0.16, and the Q-Align score is increased by 0.20-0.52.

\SubSection{Qualitative Comparison}
\begin{figure*}[t]
	\begin{center}
		\noindent
		\includegraphics[width = 4 in]{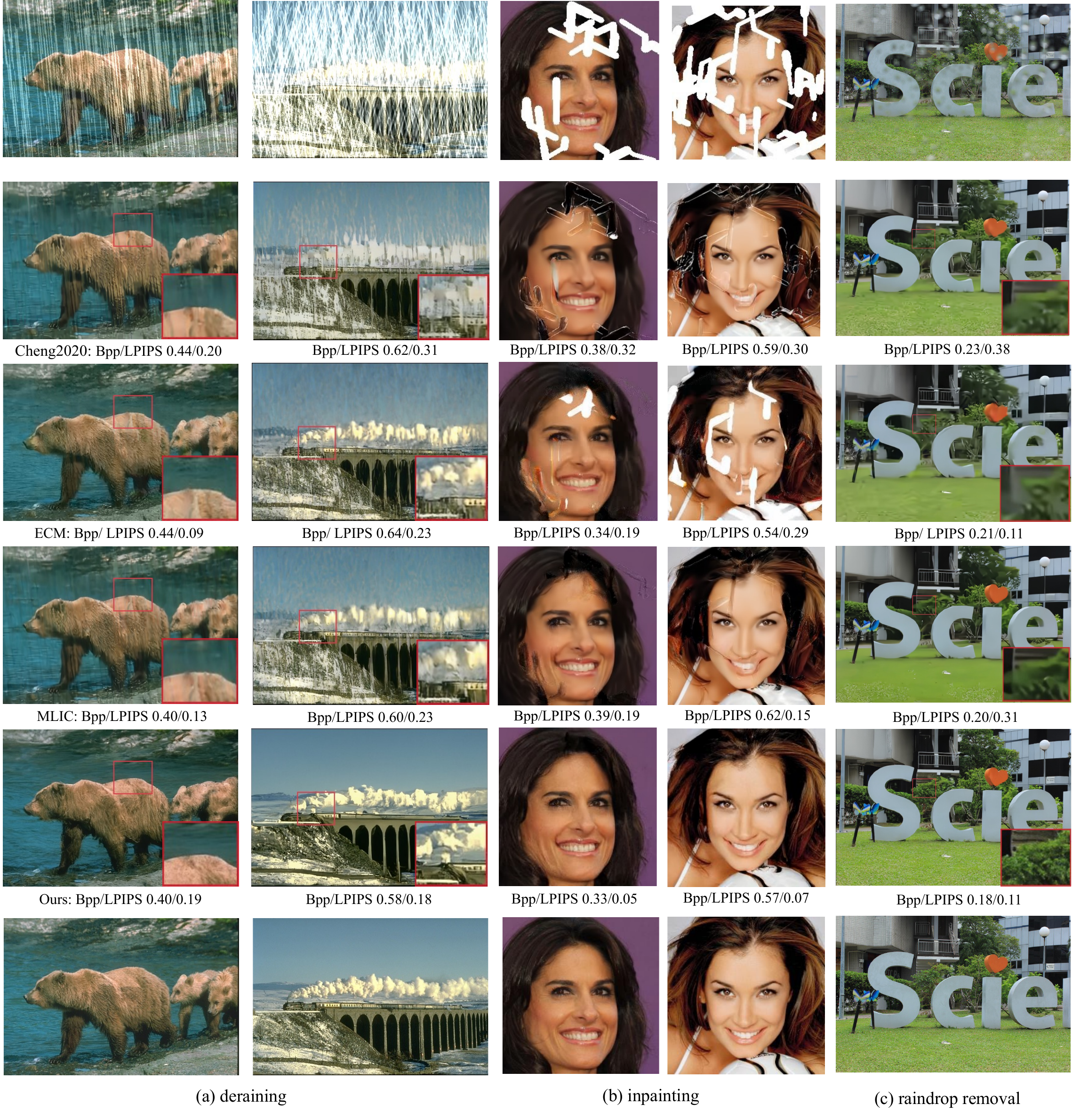}
		\caption{Qualitative comparisons between our method and other codecs on image deraining, inpainting and raindrop removal tasks. The first and the last lines show the original image $\mathbf{X}$ and the ideal high-quality image $\mathbf{X_{ideal}}$. The other lines show the enhanced image $\mathbf{\hat{X}_H}$ generated by different codecs. }\label{visual compare}
	\end{center}
\end{figure*}
Fig. \ref{visual compare} demonstrates the qualitative comparison of generated $\mathbf{\hat{X}_H}$ images by LL-ICM against the other anchors. The compressed images are in similar bit rates. Two exemplary LL tasks are showcased to demonstrate the superior performance of our LL-ICM framework, which are image deraining and raindrop removal. Obviously, our method preserves more intricate details and texture in the reconstructed and enhanced images. For example, the result for the deraining task displayed in Fig. \ref{visual compare} reveals that MLIC introduces significant rippling artifacts, while our method keeps the image content clean. Similarly, for the raindrop removal task, the results of other codecs lack details and are not clear, while our method provides the most satisfying image quality.

\Section{Conclusion}
In this paper, we propose the first image compression framework for low level machine vision tasks, LL-ICM. LL-ICM aims to increase the performance of both compression and LL vision processing, addressing that many original images to be compressed may have quality flaws and need to be enhanced by LL vision models after compression. Built upon a capable neural image codec MLIC, we incorporate a large vision-language model into the proposed LL-ICM framework, extracting generalized features to support multiple downstream LL vision tasks simultaneously. The compression and LL vision processing are then jointly optimized during training. Extensive experiment results demonstrate that our LL-ICM framework significantly outperforms existing image codecs on different LL vision tasks.

\Section{References}
\tiny
\bibliographystyle{IEEEbib}
\bibliography{refs}

\end{document}